\documentclass{article}
\usepackage{ijcai24}

\usepackage{times}
\usepackage{soul}
\usepackage[hyphens]{url}
\usepackage[hidelinks]{hyperref}
\usepackage[utf8]{inputenc}
\usepackage[small]{caption}
\usepackage{graphicx}
\usepackage{amsmath}
\usepackage{amssymb}
\usepackage{amsthm}
\usepackage{booktabs}
\usepackage[switch]{lineno}
\usepackage{multirow}

\usepackage{listings}
\usepackage{enumitem}
\usepackage{accsupp}
\usepackage{array}
\usepackage{booktabs}
\usepackage{makecell}
\usepackage{multicol}
\usepackage{xcolor}
\usepackage{todonotes}
\usepackage[utf8]{inputenc}

\usepackage{fancyhdr} %
\fancypagestyle{firstpage}
{
    \fancyfoot[C]{Accepted for presentation at the International Joint Conference on Artificial Intelligence (IJCAI), 2024}
}

\title{ScreenAI: A Vision-Language Model for UI and Infographics Understanding}

\author{
Gilles~Baechler \thanks{Equal contribution. Correspondence: jdchen@google.com}
\and
Srinivas~Sunkara \footnotemark[1]
\and
Maria~Wang \footnotemark[1]
\and
Fedir~Zubach
\and
Hassan~Mansoor
\and
\linebreak %
Vincent~Etter
\and
Victor C\u{a}rbune
\and
Jason Lin
\and
Jindong Chen \footnotemark[1]\thanks{Project leads}
\and
Abhanshu Sharma \footnotemark[2]
\affiliations
Google DeepMind
}

\begin{document}

\maketitle
\thispagestyle{firstpage}

\begin{abstract}

Screen user interfaces (UIs) and infographics, sharing similar visual language and design principles, play important roles in human communication and human-machine interaction.
We introduce ScreenAI, a vision-language model that specializes in UI and infographics understanding.
Our model improves upon the PaLI architecture with the flexible patching strategy of pix2struct and is trained on a unique mixture of datasets.
At the heart of this mixture is a novel screen annotation task in which the model has to identify the type and location of UI elements.
We use these text annotations to describe screens to Large Language Models and automatically generate question-answering (QA), UI navigation, and summarization training datasets at scale.
We run ablation studies to demonstrate the impact of these design choices.
At only 5B parameters, ScreenAI achieves new state-of-the-art results
on UI- and infographics-based tasks (Multipage DocVQA, WebSRC, and MoTIF), and new best-in-class performance on others (ChartQA, DocVQA, and InfographicVQA) compared to models of similar size.
Finally, we release three new datasets: one focused on the screen annotation task and two others focused on question answering.

\end{abstract}

\section{Introduction}

\begin{figure*}
\includegraphics[width=\textwidth,bb=0 0 720 294]{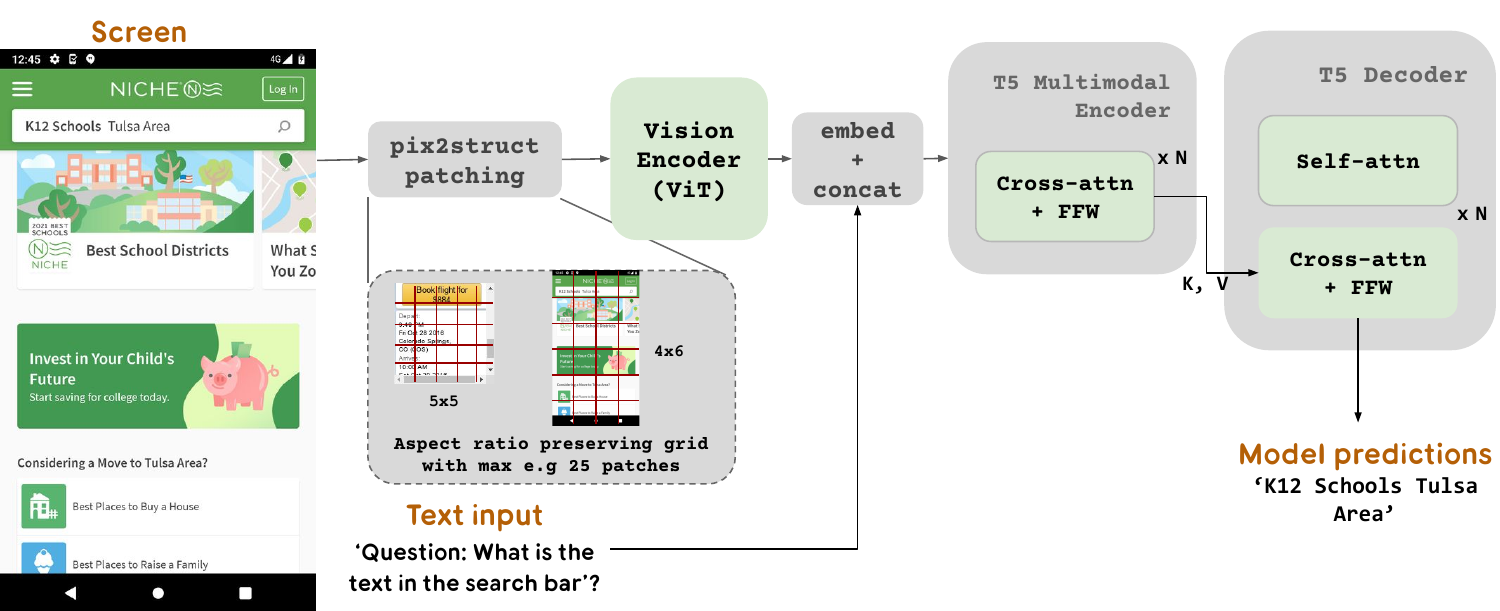}
\caption{
The overall architecture of our model. The model contains an image encoder followed by a multimodal encoder consuming embedded text and image features.
The output of the multimodal encoder is fed to an autoregressive decoder to generate the final text output.
This figure also illustrates pix2struct patching, where the grid size adapts to the aspect ratio and shape of the image.}
\label{fig:model_architecture}
\end{figure*}

Infographics, such as charts, diagrams, illustrations, maps, tables, and document layouts have long been a cornerstone of effective communication, thanks to their ability to distill complex data and ideas into simple illustrations through arrangement of layouts, and visual cues.
In the digital era, mobile and desktop UIs, sharing similar design principles and visual languages with infographics, facilitate human communication and human-machine interface with rich and interactive user experiences.

Although the above observation suggests an opportunity for a unified model,
because of their complexity, infographics and UIs present a unique challenge to
building a single model that can understand, reason, and interact on top of pictorial pixels.
To address this challenge, we introduce ScreenAI, a Vision-Language Model (VLM) for comprehensive UI and infographics understanding, including tasks
such as
question-answering (QA) on infographics (charts, illustrations, maps, etc.), and
element annotation, summarization, navigation, and QA on UIs.
Our model combines the PaLI~\cite{chen2023pali-3} architecture with the flexible patching mechanism of Pix2struct~\cite{lee2023pix2struct} and handles vision tasks by recasting them as (text, image)-to-text problems.
Figure~\ref{fig:model_architecture} provides a high level description of the model architecture and Section~\ref{sec:architecture} describes its components in more detail.

The main contributions of this work are multifold and greatly advance the field of digital content understanding:
\begin{itemize}
\item
We propose ScreenAI, a Vision-Language Model (VLM), as a holistic solution that focuses on understanding UIs and infographics, taking advantage of their common visual language and design sophistication.
\item
We introduce a textual representation for UIs, which we use to teach our model how to understand UIs during its pretraining phase.
\item
We take advantage of this new UI representation and Large Language Models (LLMs) to automatically generate training data at scale.
\item 
We define pretraining and fine-tuning mixtures which cover a wide spectrum of tasks in UI and infographic understanding.
\item
We release three evaluation datasets for tasks described in Section~\ref{sec:fine-tuning-mixtures}: Screen Annotation, ScreenQA Short, and Complex ScreenQA. These datasets enable the research community to utilize our textual representation and allow for a more comprehensive benchmarking of models for screen-based question answering.
\end{itemize}

These innovations position ScreenAI as the go-to VLM for any digital content understanding task, ranging from UIs to infographics, and beyond. At a modest size of 4.6 billion parameters, 
dated on January 17, 2024\footnote{The full paper submission deadline of IJCAI-24.},
our model exhibits state-of-the-art (SoTA) performance
on three public infographics QA benchmarks, surpassing other models 10x or more in size.
In other tasks, ScreenAI exhibits best-in-class, or close-to-best performance.
We show in Section~\ref{sec:model_size} that the model performance gets better as we increase its size, suggesting that there is a strong potential for further gains in performance by scaling up the model.

\subsection{Related Work} 
We identify three categories of closely related works.

\paragraph{Screen-Based UI Models.}
Until recently, most screen understanding efforts focused on well-defined tasks with a narrow scope. Examples include the detection of icons~\cite{zang2021multimodal} or various UI elements~\cite{zhang2021screen,sunkara2022towards,li2022learning}, together with their structure~\cite{wu2021screen}. Other notable works encompass the description of icons (widget captioning)~\cite{li2020widget}, screen summarization~\cite{wang2021screen2words}, and single-step navigation tasks~\cite{wichers2018resolving,li2022mug}.
Another direction is to use LLMs to classify and describe UI elements~\cite{gur2022understanding}, or complete tasks~\cite{nakano2021webgpt,rawles2023android,deng2023mind2web}.

\paragraph{Generalist Foundation Models.}  
The advent of large foundation models, particularly in the multimodal domain, has led to the development of versatile and unified models. These universal models excel in a broad spectrum of image understanding tasks formulated through natural language such as question-answering, image captioning, and object localization. (e.g. UniTAB~\cite{yang2022unitab}, OFA~\cite{wang2022ofa}, PaLI~\cite{chen2022pali,chen2023pali-x,chen2023pali-3}, Flamingo~\cite{alayrac2022flamingo}, or MaMMUT~\cite{kuo2023mammut}). Foundational work also includes pix2seq~\cite{chen2021pix2seq}, which recasts the object detection problem as a text prediction task.

\paragraph{Efficient Vision-Language Models.} %
Closer to the domain of screen and document understanding, similar transformer-based~\cite{vaswani2017attention} architectures have been proposed for solving various document-understanding tasks (e.g. LayoutLMv3~\cite{huang2022layoutlmv3}, Donut~\cite{kim2021donut}, pix2struct~\cite{lee2023pix2struct}, MatCha~\cite{liu2022matcha}, UDOP~\cite{tang2023unifying}, or Spotlight~\cite{li2023spotlight}). Another example is VuT~\cite{li2021vut}, which is made of a multimodal encoder, followed by a text decoder and a dedicated head for object detection tasks.

Other approaches like UIBert~\cite{bai2021uibert}, DocLLM~\cite{wang2023docllm} perform screen- and document-understanding using additional textual data extracted from metadata like DOM or ancillary models like OCR.

In our paper, we introduce pre-training tasks along with a data generation schema using self-supervision and model-based annotation.
Prior work with self-supervised learning tasks have typically been focused on one domain.
For examples, pix2struct~\cite{lee2023pix2struct}, HTLM~\cite{aghajanyan2021htlm} are focused on web-pages; ActionBert~\cite{he2021actionbert}, UIBert~\cite{bai2021uibert} are focused on mobile apps, which can capture a subset of the elements like text and exclude hierarchy information.
Our representation, inferred from only screen or image pixels, is applicable to a wide range of domains beyond web-pages and mobile apps, including documents, infographics, etc. 
Compared to prior work, our model achieves superior performance on downstream tasks. We hypothesize this is due to the positive transfer of performance when using screen, document and infographics data jointly in the pre-training mixture. Given the abundance of data in each of these domains, we believe future research in this direction can result in further improvements.

\section{Methodology}

\subsection{Architecture}\label{sec:architecture}

Our model architecture as shown in Figure~\ref{fig:model_architecture} is inspired by the architecture of the PaLI family of models~\cite{chen2022pali,chen2023pali-x,chen2023pali-3}, which is composed of a multimodal encoder block with a vision encoder like ViT~\cite{dosovitskiy2020image} and a mT5~\cite{xue2020mt5,raffel2020exploring} language encoder consuming image and text  inputs, followed by an autoregressive decoder. The input image is transformed into a sequence of embeddings by the vision encoder and these embeddings are concatenated with the input text embeddings and fed into the mT5 language encoder. The output of this encoder is passed to the decoder to generate the text output. This generic formulation enables us to use the same model architecture to solve a variety of vision and multimodal tasks that can be recast as a text+image (input) to text (output) problem. Compared to the text input, the image embeddings constitute a significant portion of the input length to the multimodal encoder.

We further extend PaLI’s encoder-decoder architecture to accept various image patching patterns. The original PaLI architecture only accepts a fixed grid pattern of patches for processing the input images.
However, the data we encounter in screen-related domains spans a wide variety of resolutions and aspect ratios. To have a single model to work across all screen shapes, it is necessary to use a patching strategy which can work well with images of various shapes. To this end, we borrow a technique introduced in Pix2Struct~\cite{lee2023pix2struct}, which allows us to have image patches with arbitrary grid shapes based on the input image shape and a pre-defined maximum number of patches, as shown in Figure~\ref{fig:model_architecture}. 
This enables us to accommodate input images of various formats and aspect ratios without the need for padding or stretching the image to a fixed shape, making our model more polyvalent to handle both mobile (i.e. portrait) and desktop (i.e. landscape) image formats. In Section~\ref{experiments-section}, we evaluate the impact of each of these modeling choices.

\subsection{Model Configurations}
\label{sec:model_configs}
We train models of 3 different sizes containing 670M, 2B and 5B parameters. For the 670M and 2B parameter models, we start from pre-trained unimodal checkpoints for the vision encoder and the encoder-decoder language models. For the 5B parameter model, we start from the multimodal pre-trained checkpoint from PaLI-3~\cite{chen2023pali-x}, where the ViT is trained together with the UL2~\cite{tay2022ul2} based encoder-decoder language model. 
A breakdown of the parameter distribution among the vision and language models can be seen in Table~\ref{tab:models-sizes}. 

Our patching strategy allows variable aspect ratios and input resolutions, as long as they fit within the allocated sequence length budget ($2024$ embeddings for the 670M model, $2916$ embeddings for the 2B model, and $3364$ embeddings for the 5B model). 
For square images, the corresponding maximum input resolution is $720 \times 720$ for the 670M model, $756 \times 756$ for the 2B model, and $812 \times 812$ for the 5B model.

\begin{table}
\centering
\begin{tabular}{l l l r}
		\toprule
\textbf{Model} & \textbf{ViT} & \textbf{Encoder-Decoder} & \textbf{\#params}  \\
\midrule
670M & B16 ($92 \text{M}$) & mT5 base ($583 \text{M}$) & $675 \text{M}$ \\
2B & H14 ($653 \text{M}$) & mT5 Large ($1.23 \text{B}$) & $1.88 \text{B}$ \\
5B & G14 ($1.69 \text{B}$) & UL2-3B ($2.93 \text{B}$) & $4.62 \text{B}$ \\
\bottomrule
\end{tabular}
\caption{Model variants and details of their parameter counts and split among vision and language models. The image encoders are based on ViT~\protect\cite{dosovitskiy2020image} and the text encoders are based on mT5~\protect\cite{xue2020mt5} and UL2 models~\protect\cite{tay2022ul2}.
}
\label{tab:models-sizes}
\end{table}

\subsection{Stages of Training}
In this section, we cover the different stages of training.

\paragraph{Pre-Training.}

Starting from the checkpoints mentioned in Section~\ref{sec:model_configs}, we do a first stage of training on large datasets generated from self-supervision and other models, using minimal human labeling (see Section~\ref{sec:pretraining-mixtures} for a detailed description of the pre-training mixture). 
Contrary to the later fine-tuning stage, we train both the vision encoder and the language model.
The motivation behind training the vision encoder is to incorporate the new patching strategy, and to allow the model to adapt from natural images to UI-related images.
We evaluate the impact of training the vision encoder and of including LLM generated data on a variety of tasks in our ablation experiments in Section~\ref{experiments-section}.

After some initial steps of pretraining, we perform additional steps with the ViT encoder frozen to further train the model while reducing the resource consumption.

\paragraph{Fine-Tuning.}
During fine-tuning, the model is trained on mixtures of tasks, most of which are labeled using human annotators. These tasks are described in details in Section~\ref{sec:fine-tuning-mixtures}.
For QA-related tasks, we start by fine-tuning the model on a combination of QA-related tasks; then, additional training is performed on each individual tasks separately.
For all other tasks, we fine-tune the model on each one individually.

\section{Automatic Data Generation}\label{sec:automatic-data-gen}

The pretraining phase of our model's development is critically dependent on access to a vast and diverse dataset. Given the impracticality of manually annotating such an extensive dataset, our strategy focuses on automatic data generation. This approach leverages specialized smaller models, each adept at generating and labeling data both efficiently and with a high degree of accuracy.

In this section, we provide a detailed account of our data generation process, particularly highlighting how we gather and automatically annotate a diverse range of screenshots for pretraining our model. This automated approach is not only efficient and scalable compared to manual annotation but also ensures a level of data diversity and complexity.

\subsection{Screen Annotation}
\label{sec:screen_annotation}

\begin{figure*}
\includegraphics[width=\textwidth]{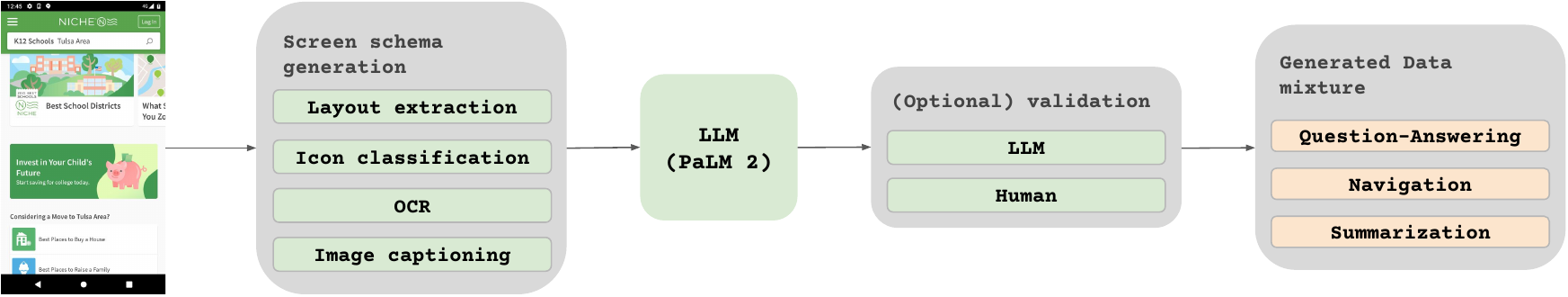}
\caption{Task generation pipeline: 1) the screens are first annotated using various models; 2) we then use an LLMs to generate screen-related tasks at scale; 3) (optionally) we validate the data using another LLM or human raters.}
\label{fig:llm_generated_data}
\end{figure*}

Our initial step is to equip the model with a comprehensive understanding of textual elements, various screen components, and their overall structure and hierarchy.
This foundational understanding is vital for the model's ability to interpret and interact accurately with a wide range of user interfaces.

An extensive collection of screenshots has been amassed from various devices, including desktops, mobile, and tablets, by crawling applications and web pages~\cite{raffel2020exploring}. These screenshots are then annotated with detailed labels that describe the UI elements, their spatial relationships, and additional descriptive information.

The cornerstone of our annotation process is a layout annotator based on the DETR~\cite{carion2020end} detection model. This object detector is apt at identifying and labeling a wide range of UI elements such as IMAGE, PICTOGRAM, BUTTON, TEXT, and others. This detector and the list of UI elements is inspired by~\cite{li2022learning}.
However, 
the models in~\cite{li2022learning} are classifiers and are provided a list of candidate bounding boxes to annotate, whereas in our case we predict the bounding boxes too.

Pictograms undergo further analysis using an icon classifier~\cite{sunkara2022towards} capable of distinguishing 77 different icon types. This detailed classification is essential for interpreting the subtle communication conveyed through icons.
For icons that are not covered by the classifier, infographics and images, we use the PaLI image captioning model~\cite{chen2023pali-3}.
This model generates descriptive captions that provide contextual information, aiding in the comprehensive understanding of the screen's content.

Additionally, an OCR engine extracts and annotates textual content on screen. This step is crucial for interpreting the textual information presented in various formats on interfaces.
Finally, we combine the OCR text with the previous annotations to create a detailed and holistic description of each screen. The bounding box coordinates are systematically included, providing spatial context to the elements on the screen.

\begin{figure}
\includegraphics[scale=0.28]{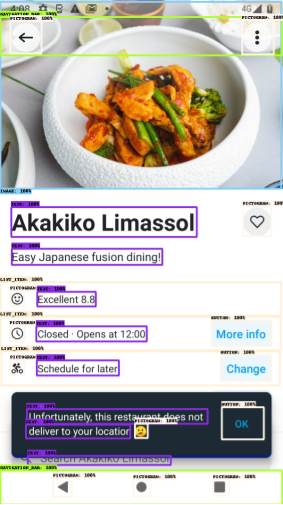}
\includegraphics[scale=0.28]{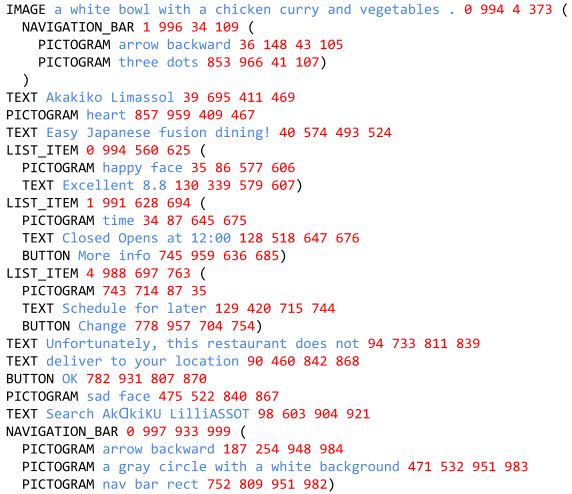}

\caption{Example of our screen schema. See
Appendix~\ref{app:screen_schema}
for more.
}
\label{fig:screen_schema_p2}
\end{figure}

Figure~\ref{fig:screen_schema_p2} shows an example of the screen schema used in most of our pretraining tasks. Each schema contains:
\begin{enumerate}
    \item The UI element names.
    \item The OCR text (when applicable).
    \item The element descriptions, e.g. captioning or icon names.
    \item The bounding box coordinates, quantized and normalized between $0$ and $999$.
\end{enumerate}
Parentheses are used to create a basic hierarchical structure between the elements, i.e. the children of a parent element are all put inside a parenthesis block. For ease of visualization, the bounding boxes from the screen schema have been overlaid on the original screenshot.

This schema plays a central role in our data generation for pretraining tasks, offering a detailed and multifaceted representation of screen content. The schema itself also serves as a pretraining task, where the model is tasked with generating a similar schema from a provided input image. This not only enhances the model's capacity to discern and interpret various UI components but also their relationships to one another.
Additionally, the screen schema proves to be an invaluable natural language tool to interface with large language models (LLMs). By providing LLMs with a structured and detailed representation of screen content, we enable the creation of more intricate and contextually nuanced tasks.

\subsection{LLMs to Generate Additional Tasks}

To infuse greater diversity into our pretraining data, we leverage the capabilities of LLMs, in particular PaLM 2-S~\cite{anil2023palm} to generate Question-Answer pairs in two stages.
Initially, we generate the screen schema as previously described. Subsequently, we craft a prompt incorporating the screen schema and direct the LLM to generate synthetic data.
This stage is empirical and necessitates a degree of prompt engineering. However, after several iterations, we typically identify a prompt that effectively generates the desired task. Example of such prompts are shown in
Appendix~\ref{app:qa_example_prompt}.
To evaluate the quality of these generated responses, we conducted human validation on a subset of the data, ensuring that it meets a predetermined quality threshold.

This approach is described in Figure~\ref{fig:llm_generated_data} and it enables us to create a variety of synthetic but realistic tasks that significantly enhance the depth and breadth of our pretraining dataset. By leveraging the natural language processing capabilities of LLMs, coupled with the structured screen schema, we can simulate a wide range of user interactions and scenarios. See
Appendix~\ref{appendix:screen_navigation_generated_examples}
for generated examples.

\section{Data Mixtures}

We define two distinct sets of tasks for our model: an initial series of pretraining tasks and a subsequent set of fine-tuning tasks.
The distinction primarily lies in two aspects:
\begin{enumerate}
    \item 
    \textbf{Source of the Groundtruth Data:} For the fine-tuning tasks, the labels are provided or verified by human raters. For the pretraining tasks, the labels are inferred using self supervised learning methods or generated using other models.
    \item
    \textbf{Size of the Datasets:} Typically, the pretraining tasks encompass a significantly larger quantity of samples, and consequently, these tasks are used for training the model over a more extended series of steps.
\end{enumerate}

\subsection{Pretraining Mixture}\label{sec:pretraining-mixtures}

\begin{figure*}[t]
\centering
\includegraphics[width=.9\textwidth]{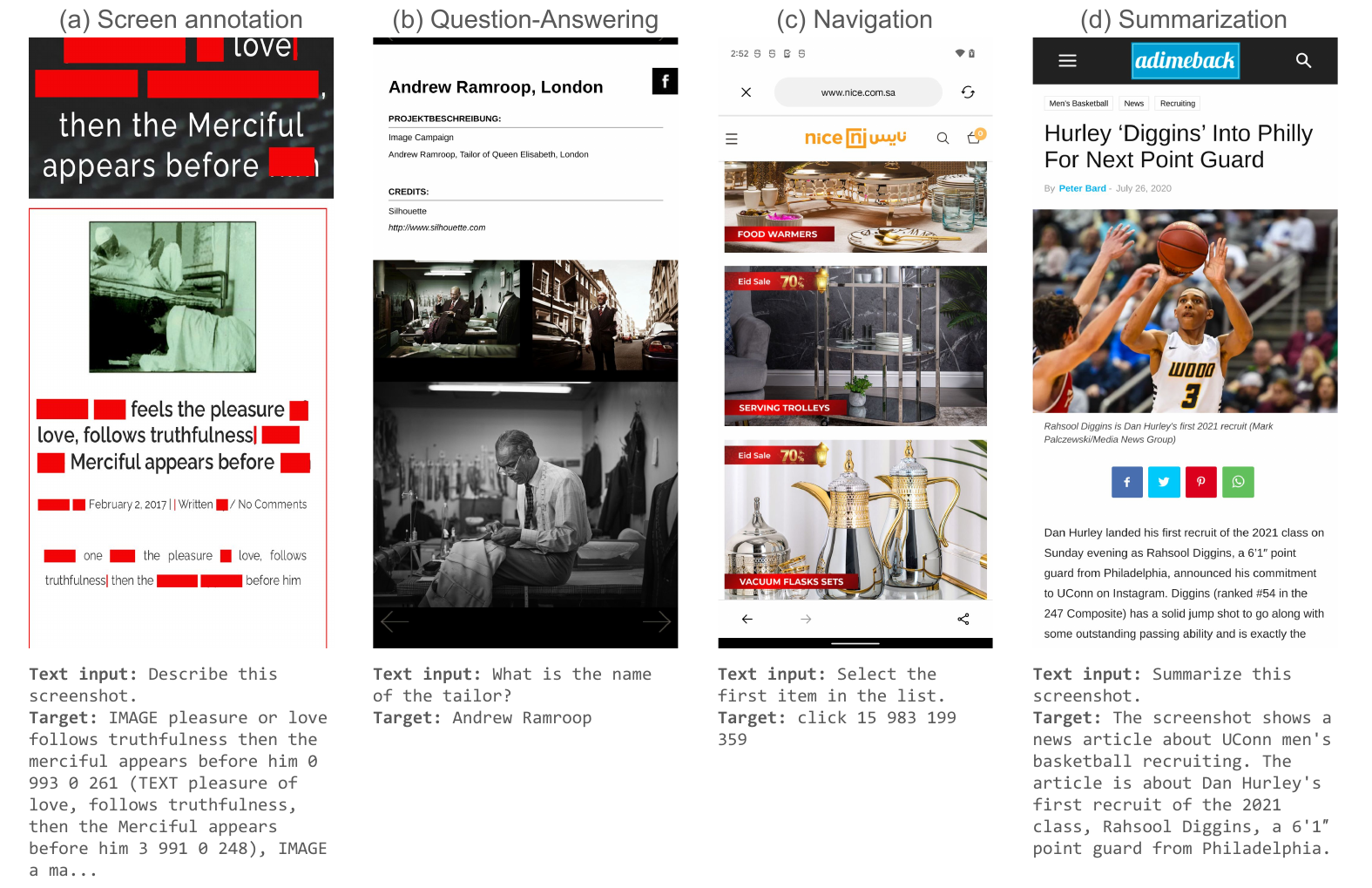}
\caption{Sample of tasks that we are using in our pretraining mixture: (a) Screen annotation, with masking; (b) Question-Answering; (c) Navigation; (d) Summarization. The last three have been generated using our screen annotation model, coupled with PaLM-2-S.}
\label{fig:pretraining_mixture}
\end{figure*}

Based on the methodology outlined in Section~\ref{sec:automatic-data-gen}, we have selected the following tasks for pretraining our models. These tasks, each illustrated in Figure~\ref{fig:pretraining_mixture}, are designed to cover a wide range of skills and scenarios, endowing our model with diverse real-world applications.

\begin{enumerate}
\item
\textbf{Screen Annotation:}
The model is tasked with detecting and identifying UI elements present on a screen. This includes performing OCR and image captioning to understand and interpret the textual and non-textual content. To enhance the model's contextual understanding, some text elements are intentionally masked, encouraging the model to infer information based on the surrounding context and layout.

\item
\textbf{Screen Question-Answering (QA):}
For this task, the model is asked to answer questions related to user interfaces and computer-generated images, such as infographics. After initial experiments, we identified certain gaps in performance on attributes like arithmetic, counting, understanding images with complex infographics. To enhance the model capabilities, we create data specifically addressing these gaps, e.g., QA involving counting, arithmetic operations, and complex data containing infographics. For these examples, we first crawl large scale webpage and infographic images, then perform prompt tuning to generate and validate relevant questions and their answers. For charts, the mix consists of 1) synthetic data \cite{liu2023deplot}, 2) UniChart \cite{masry2023unichart}, 3) DVQA \cite{kafle2018dvqa}, 4) TaTa \cite{gehrmann2022tata}, 5) Benetech \footnote{\url{https://www.kaggle.com/competitions/benetech-making-graphs-accessible}}.

\item
\textbf{Screen Navigation:} This task involves interpreting navigation instructions (e.g., `go back') and identifying the appropriate UI element to interact with. The expected output is the bounding box coordinates of the target element, bucketized between $0$ and $999$, demonstrating the model's ability to understand user intent and navigate through interfaces accurately.

\item
\textbf{Screen Summarization:}
The model is tasked to succinctly summarize the content of a screen in one or two sentences. This task assesses the model's capability to distill and caption the essence of the screen's content.

\end{enumerate}

To ensure comprehensive training robust to aspect ratios, each task is made available across multiple formats (mobile and desktop) and includes several aspect ratios. 

\begin{table}
\centering
\begin{tabular}{lr}
\toprule
\textbf{Task Name}  & \textbf{\#samples}\\
\midrule
\textbf{Generated Screen Annotation} \\
Mobile webpages & $262 \text{M}$\\ 
Mobile apps & $54 \text{M}$ \\
Mobile webpages (tall renders) & $37\text{M}$\vspace{1mm}\\ 
\textbf{Generated Screen Question-Answering} \\
Mobile webpages & $9.8\text{M}$\\ %
Mobile apps & $2.0\text{M}$\\ %
Mobile webpages (tall renders) & $2.3\text{M}$\\
Desktop webpages &  $16.4\text{M}$\\ %
Infographics & $6.3\text{M}$\\ %
ChartQA/PlotQA & $2.4\text{M}$\vspace{1mm}\\
\textbf{Generated Screen Navigation} \\
Mobile webpages & $2.6 \text{M}$\\
Mobile apps & $5.9 \text{M}$ \\
Mobile webpages (tall renders) & $2.3\text{M}$\\
Desktop webpages & $5.1\text{M}$\vspace{1mm}\\ 
\textbf{Generated Screen Summarization} \\
Mobile webpages & $5.6\text{M}$\\
Desktop webpages & $7.6\text{M}$\vspace{1mm}\\
\textbf{Other} \\
Tarzan~\cite{xue2020mt5} & $297 \text{K}$ \\
VQA CC3M~\cite{sharma2018conceptual} & $178 \text{K}$ \\
WebLI Alt and OCR text~\cite{kil2023prestu} & $297 \text{K}$\\ 
\bottomrule
\end{tabular}
\caption{Detailed breakdown of our pretraining mixture.}
\label{tab:pretraining_mixture}
\end{table}

In addition to these screen-related tasks, our training regimen also incorporates a variety of other image and text data sources: Span corruption on C4~\cite{xue2020mt5}, VQA CC3M~\cite{sharma2018conceptual}, WebLI Alt and OCR text~\cite{kil2023prestu,chen2022pali} and Chart-to-table translation~\cite{liu2023deplot}.
Such datasets have been instrumental in the development of PaLI models~\cite{chen2022pali,chen2023pali-3}, which serve as the foundational architecture for our model.
Their inclusion ensures that our model not only excels in screen and infographics understanding but also maintains robust language and visual processing capabilities.

A summary of all our pretraining tasks is shown in Table~\ref{tab:pretraining_mixture}.
In the mixture, datasets are weighted proportionally to their size with a maximum allowed weight per task.
Incorporating multimodal sources in our multi-task training, from language processing to visual comprehension and web content analysis, prepares our model to handle diverse scenarios effectively and enhances its overall versatility and performance.

\subsection{Fine-Tuning Tasks and Benchmarks}
\label{sec:fine-tuning-mixtures}

We use a variety of tasks and benchmarks during fine-tuning to estimate the quality of our model.
These benchmarks are summarized in Table~\ref{tab:finetuning_mixture} and include the main existing screen, infographics and document understanding benchmarks.
We make the following changes to task formulations:
(1) we cast RefExp~\cite{wichers2018resolving} and Task Automation in MoTIF~\cite{burns2022motif} as object detection tasks, without using candidate bounding boxes and report accuracy at IoU=0.1\footnote{Intersection over union at threshold 0.1}
considering only one box predicted;
(2) for MoTIF, we report the number for the app-unseen split of the test set in Table~\ref{tab:results}, and other split results in
in Table \ref{tab:motif_results_deails} of Appendix~\ref{appendix:motif_eval_results}.

\begin{table}[t]
\centering
\begin{tabular}{ll}
\toprule
\textbf{Task Name/Benchmark}  & \textbf{Metric} \\
\midrule
\textbf{Screen Analysis} \\
Screen Annotation [Ours, Sec.~\ref{sec:screen_annotation_benchmark}] & F1@IoU=0.1 \\
Widget Captioning~\cite{li2020widget} & CIDEr\vspace{1mm}\\
\textbf{Screen Question-Answering} \\
ScreenQA Short [Ours, Sec.~\ref{sec:screen_qa_short_benchmark}] & SQuAD F1\\
Complex ScreenQA [Ours, Sec.~\ref{sec:complex_qa_benchmark}] & SQuAD F1\\
WebSRC~\cite{chen2021websrc} & SQuAD F1 \vspace{1mm}\\
\textbf{Screen Navigation} \\
RefExp~\cite{bai2021uibert} & Acc@IoU=0.1 \\
MoTIF-Automation~\cite{burns2022motif} & Acc@IoU=0.1\vspace{1mm}\\
\textbf{Screen Summarization} \\
Screen2Words~\cite{wang2021screen2words} & CIDEr\vspace{1mm}\\
\textbf{Infographics/Doc Visual QAs} \\
ChartQA~\cite{masry2022chartqa} & Relaxed Acc.\\
DocVQA~\cite{mathew2021docvqa} & ANLS\\
Multipage DocVQA~\cite{tito2023hierarchical} & ANLS\\
InfographicVQA~\cite{mathew2022infographicvqa} & ANLS\\
OCR-VQA-200K~\cite{mishraICDAR19} & Exact Match\\
\bottomrule
\end{tabular}
\caption{
Detailed breakdown of our fine-tuning mixture and their associated metrics.
We assume readers are familiar with these metrics, but include descriptions and citations in 
Appendix~\ref{app:metrics}
for reference.
}
\label{tab:finetuning_mixture}
\end{table}

\newcommand{\sota}[2]{\makecell[l]{$#1^{#2}$}}
\newcommand{\na}{\makecell[c]{-}}

\begin{table*}
\centering
\begin{tabular}
{llllllllllllll}
\toprule
& \thead{SA}  & \thead{Ref \\ Exp} & \thead{SQA \\Short} & \thead{Cplx \\ SQA} & \thead{\small{MoTIF}} & \thead{Screen2 \\ Words} & \thead{Widget \\ Capt.} & \thead{Chart \\ QA} & \thead{Doc \\ VQA} & \thead{MPDoc \\ VQA} & \thead{Info \\ VQA} & \thead{OCR \\ VQA} & \thead{Web \\ SRC} \\

\midrule
SoTA & \na & \na & \na & \na & \sota{67.6}{a}  & \sota{\textbf{130.7}}{b} & \sota{\textbf{159.8}}{b} & \sota{\textbf{80.8}}{h} & \sota{\textbf{90.9}}{h} & \sota{61.8}{d} & \sota{\textbf{80.3}}{h} & \sota{\textbf{77.8}}{b} & \sota{85.0}{f} \\
\midrule
\multicolumn{14}{l}{\small \textbf{Without OCR}} \\
\small{SoTA$\leq$5B}  & \na & \na & \na & \na & \sota{67.6}{a}  & \sota{130.7}{b} & \sota{159.8}{b} & \sota{\underline{77.3}}{i} & \sota{\underline{87.8}}{c} & \na & \sota{57.8}{b} & \sota{\underline{76.7}}{b} & \sota{77.8}{g} \\
ScreenAI & \textbf{86.2} & \textbf{86.3} &  \textbf{94.6} & \textbf{42.4} & \textbf{87.4}  & 120.8 & 156.4 & 76.6 & 87.5 & \textbf{72.9} & \underline{61.4} & 75.0 & \textbf{87.2} \\
\midrule
\multicolumn{14}{l}{\small \textbf{With OCR}} \\
\small{SoTA$\leq$5B} & \na & \na & \na & \na & \na & \na & \na & \sota{70.4}{c} & \sota{89.3}{c} & \sota{61.8}{d} & \sota{62.4}{b} & \sota{\underline{77.8}}{b} & \sota{85.0}{f} \\
ScreenAI & \na & \na & \textbf{94.8} & \textbf{43.5} & \na & 123.7 & \na & \underline{76.7} & \underline{89.9} & \textbf{77.1} & \underline{65.9} & 76.2 & \na \\
\bottomrule
\end{tabular} 
\caption{Comparison of ScreenAI with various SoTA models: (a) MoTIF~\protect\cite{burns2022motif}, (b) PaLI-3~\protect\cite{chen2023pali-3}, (c) SmoLA PaLI-X~\protect\cite{wu2023omnismola}, (d) Hi-VT5~\protect\cite{tito2023hierarchical}, (e) TILT~\protect\cite{powalski2021going}, (f) DocPrompt~\protect\cite{wu2023docprompt}, (g) DUBLIN~\protect\cite{aggarwal2023dublin}, (h) Gemini~\protect\cite{geminiteam2023gemini}, (i) ChartPaLI-5B~\protect\cite{carbune2024chartbased}. 
Bold font highlights SoTA score, and underscore represents best-in-class score.
See Table~\ref{tab:finetuning_mixture} for details about the tasks and their associated metrics.}
\label{tab:results}
\end{table*}

\begin{figure}[t]
\centering
\addtolength{\tabcolsep}{-.45em}
\begin{tabular*}{.5\textwidth}{l|l|l}
\includegraphics[width=0.15\textwidth]{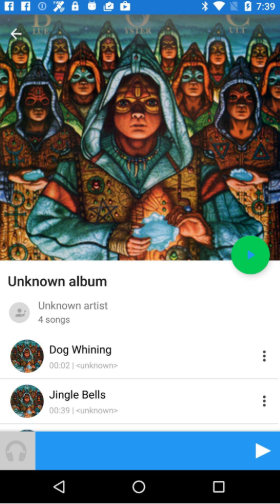} &
\includegraphics[width=0.15\textwidth]{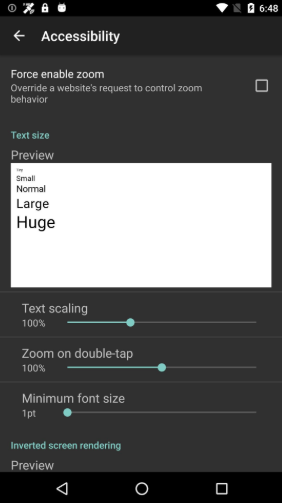} &
\includegraphics[width=0.15\textwidth]{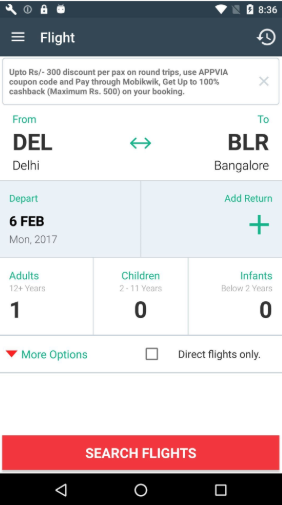} \\
\parbox{0.15\textwidth}{
\scriptsize
\textsf{\noindent
\textbf{Question:} How many songs have a duration of less than 30 seconds?\\ 
\textbf{Answer:} 1}
} &
\parbox{0.15\textwidth}{
\scriptsize
\textsf{\noindent
\textbf{Question:} How many text size options are there?\\
\textbf{Answer:} 5}
} &
\parbox{0.16\textwidth}{
\scriptsize
\textsf{\noindent
\textbf{Question:} How many days are between the departure and return dates?\\
\textbf{Answer:} There is no answer on the screen.}
}
\end{tabular*}
\caption{Examples of mobile screen in Complex QA dataset.}
\label{fig:complex_qa}
\end{figure}

\begin{figure}[t]
\centering
\includegraphics[width=0.45\textwidth]{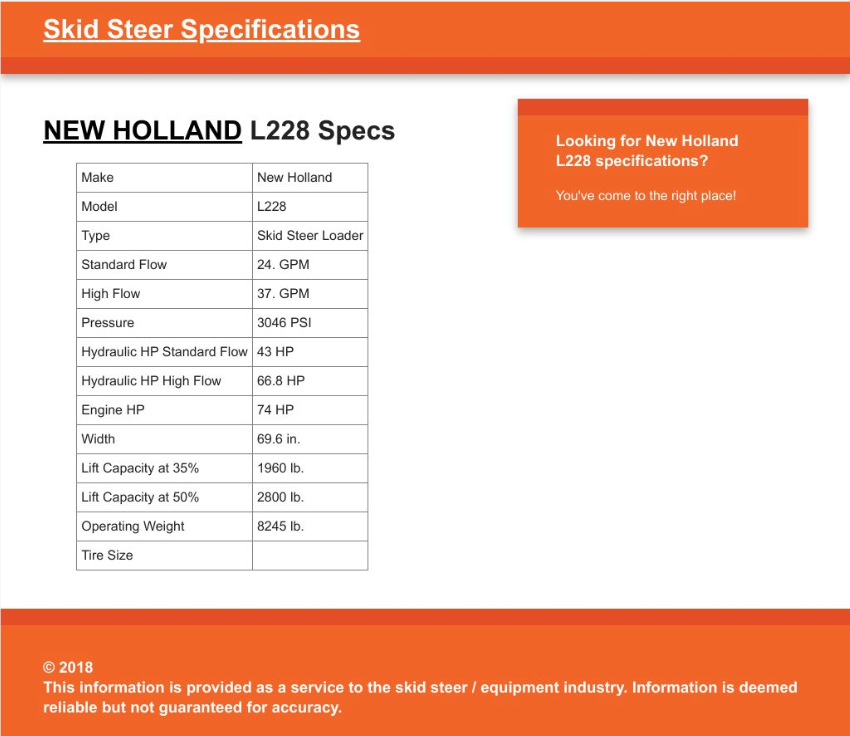}\\
\parbox{0.45\textwidth}{
\scriptsize
\textsf{
\textbf{Question:} What is the lift capacity at 35\%?
\textbf{Answer:} 1960 lb.}
}
\caption{An example of desktop screen in Complex QA dataset.}
\label{fig:complex_qa_desktop_p1}
\end{figure}

We supplement the tasks mentioned above with three new benchmarks that we release:
\begin{itemize}
    \item
    \label{sec:screen_annotation_benchmark}
    \textbf{Screen Annotation (SA):}\footnote{\url{https://github.com/google-research-datasets/screen_annotation}}
    To evaluate our model's layout annotation and spatial understanding capabilities, we create a dedicated benchmark consisting of 4.2K screenshots from the Rico dataset~\cite{deka2017rico}. Each UI element has been annotated by human raters, and
    the annotations comprise a bounding box and a UI class from the list described in~\ref{sec:screen_annotation}.
    We evaluate the model's predictions using object detection metrics, including F1 score, precision and recall values computed at IoU=0.1.
    \item
    \label{sec:screen_qa_short_benchmark}
    \textbf{ScreenQA Short (SQA Short):}\footnote{\url{https://github.com/google-research-datasets/screen_qa?tab=readme-ov-file\#screenqa-short}}
    ScreenQA~\cite{hsiao2022screenqa}, a benchmark for screen understanding,
    contains UI elements and full-sentence answers as ground truth. To align the output format with other question answering tasks, we generate a new ground truth, a list of alternative short answers, for each of the questions.
    We use the maximum F1 score across all the candidate answers as the metric.
    See Figure~\ref{fig:screenqa_short_examples_p3} and
        Appendix~\ref{appendix:screenqa_short}
    for more details.
    \item
    \label{sec:complex_qa_benchmark}
    \textbf{Complex ScreenQA (Cplx SQA):}\footnote{\url{https://github.com/google-research-datasets/screen_qa?tab=readme-ov-file\#complexqa}}
    To complement SQA Short, we introduce Complex ScreenQA, which includes more difficult questions (counting, arithmetic, comparison, and non-answerable questions) and contains screens with various aspect ratios. See
    Figures~\ref{fig:complex_qa} and~\ref{fig:complex_qa_desktop_p1} for examples and
        Appendix~\ref{appendix:complex_qa_datasets}
    for more details.
\end{itemize}

\begin{figure}[t]
\centering
\addtolength{\tabcolsep}{-.45em}
\begin{tabular*}{.5\textwidth}{l|l}
    \includegraphics[width=.15\textwidth]{screenshots/rico_app.honestly_0_457.png}
&
\begin{minipage}[b]{0.32\textwidth}
\setlength{\parskip}{0pt}
\scriptsize
\textsf{\textbf{Question:}
How many links and comments are there of the post "Why Michael Flynn kept his Job 17 days after the White House!" ?
}

\textsf{\textbf{Full sentence answers:}
\begin{itemize}[nolistsep]
    \item There is 1 like and 1 comment on the post "Why Michael Flynn kept his job 17 days after the White House!".
    \item There is 1 like and 1 comment on the "Why Michael Flynn kept his Job 17 days after the White House!" post.
    \item There is 1 like and 1 comment.
\end{itemize}
\textsf{\textbf{List of short answers:}
\begin{itemize}[nolistsep]
    \item one and one
    \item 1 and 1
    \item one, one
    \item 1, 1
    \item 1 like, 1 comment
    \item 1 like and 1 comment
\end{itemize}
}
}
\end{minipage}
\end{tabular*}
\caption{Examples of questions and answers from the ScreenQA dataset, together with their LLM-generated short answers.}
\label{fig:screenqa_short_examples_p3}
\end{figure}

We also provide a few additional details on how we handle Multipage DocVQA and ChartQA.

\paragraph{Multipage DocVQA.}
\label{sec:mpdocvqa}

The standard fine-tuning task for Multipage DocVQA \cite{tito2023hierarchical} can be transformed into a single-page DocVQA task by pairing the same question with each page of the document and choosing the answer with the highest score among all pages. In this formulation, we modify the training set by splitting a question, answer and multipage document into a positive pair (with the actual answer for the page containing the answer) and multiple negative pairs (with “no answer” for pages which do not contain the answer). The negative pairs are subsampled to avoid overfitting on not predicting an answer and the original DocVQA task \cite{mathew2021docvqa} is added to the fine-tuning mixture.

\paragraph{ChartQA.}
\label{sec:chartqa}

Concurrent work in \cite{carbune2024chartbased} showed that the original fine-tuning dataset~\cite{masry2022chartqa} is insufficiently rich for learning solving complex reasoning tasks. There, they overcome this limitation through synthetic examples and rationales, paired with training loss changes. Here, we leverage the synthetic examples, but without modifying the training loss or incorporating rationales. We therefore maintain parity how we fine-tune for the rest of the tasks. We report similar performance with or without OCR, hinting that the scale of the dataset contributes more than the input features. Our results otherwise further strengthen the contribution of the pre-training and architecture changes with pix2struct to better leverage the same synthetic examples and not needing to rely on rationales.

\begin{figure*}
    \includegraphics[width=\textwidth]{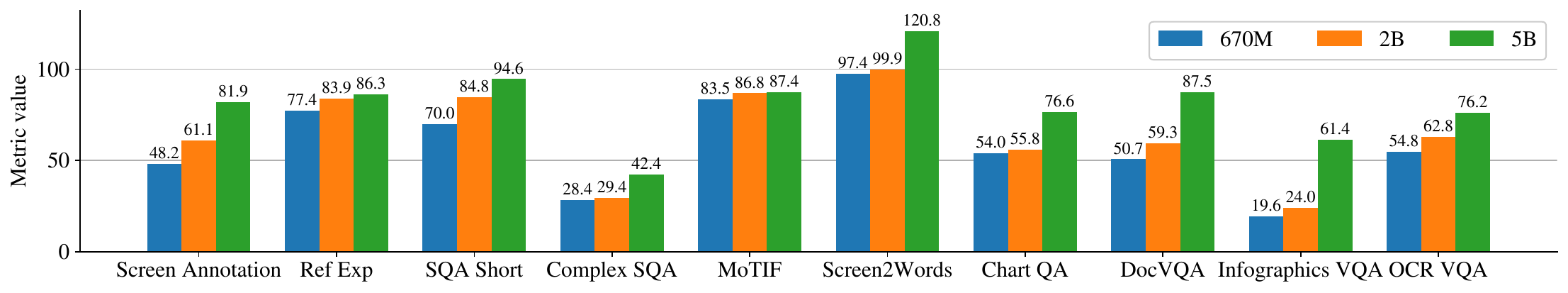}
    \caption{Performance of different model sizes on fine-tuning tasks. The metrics improve consistently as the model size increases.}
    \label{fig:results-model-size}
\end{figure*}

\section{Experiments and Results} \label{experiments-section}

In this section, we present the setup we used to conduct our experiments and analyze our findings. First, we compare the best performing ScreenAI model to the SoTA on a variety of Screen and Infographics related tasks. Next, we report the impact of model size on overall performance.
Finally, we report results on ablation studies to validate the design choices made for the models. 

\subsection{Experiments Setup}
In the fine-tuning phase, we hold the ViT encoder frozen and fine-tune the language model only. We use 512 as our batch size for fine-tuning. Our text input sequence length is 128 and output sequence length varies depending on individual tasks. When fine-tuning with OCR as additional input, we increase the input sequence length accordingly. We generally find that the model converges within 30k steps. Unless specified otherwise, all experiments are run on the 5B model.

\subsection{Results}

Table~\ref{tab:results} shows the performance of our models and compares them with state-of-the-art (SoTA) results on a variety of screen- and infographics-related tasks.
We also include the best results for models of similar size (SoTA$<$5B).
We report new SoTA results on MoTIF, MPDocVQA, and WebSRC; and new best-in-class results in ChartQA, DocVQA and InfographicVQA (InfoVQA).
We report same or competitive performance on Screen2Words, Widget Captioning, and OCR-VQA.
We also report our results on the benchmarks introduced in Section~\ref{sec:fine-tuning-mixtures} (Screen Annotations, Referring Expressions, ScreenQA Short and Complex ScreenQA).

\paragraph
{Adding OCR as Additional Input.}
We analyze the impact of adding OCR\footnote{We use a proprietary OCR system similar to GCP Vision API to produce additional OCR input for each image.} to the model input by conducting experiments with and without OCR.
This is inspired by fine-tuning experiments in PaLI~\cite{chen2023pali-3}, where across all screen- and document-related tasks, passing OCR texts as additional input improves task performance.
In Table~\ref{tab:results} we present our single task fine-tuning results using OCR data.
For QA tasks, OCR input provides a boost in performance (e.g. up to $~4.5\%$ on Complex ScreenQA, MPDocVQA and InfoVQA).
However, using OCR imposes a slightly larger input length and hence results in slower overall training.
It also requires having OCR results available at inference time.

\paragraph
{Model Size.}\label{sec:model_size}
We conducted single task experiments with the following model sizes: $670 \text{M}$, $2 \text{B}$ and $5 \text{B}$. We use benchmarks for screen tasks  as well as other public tasks. In Figure~\ref{fig:results-model-size}, we observe that across all tasks, increasing the model size improves performances and the improvements have not saturated at the largest size. We observe that for tasks that require more complex visual-text and arithmetic reasoning e.g. InfoVQA, ChartQA, and Complex ScreenQA, 
the improvement between 2B and 5B models is significantly larger than between 670M and 2B models.

\begin{figure*}
\centering
\includegraphics[width=0.8\textwidth]{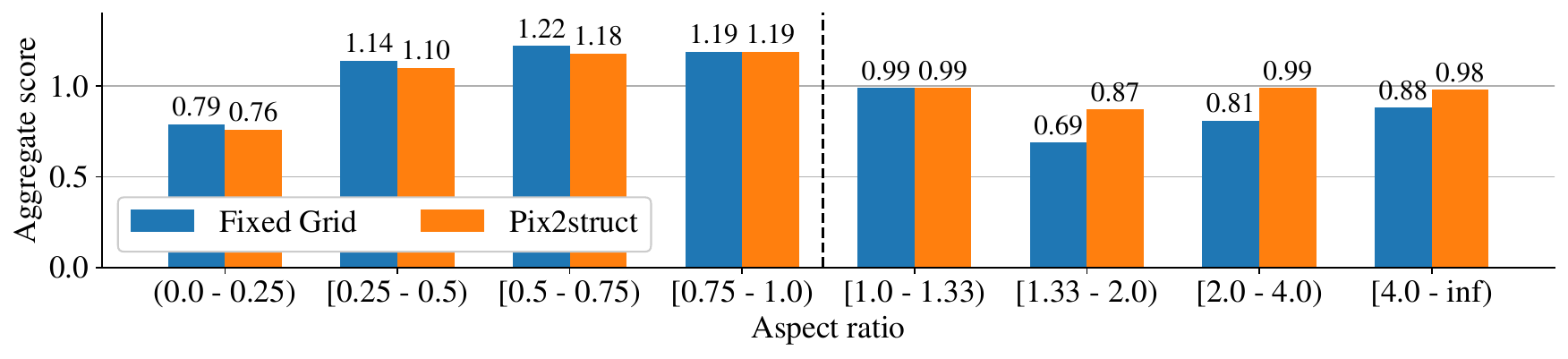}
\caption{Ablation study for Pix2Struct vs. fixed-grid patching; the numbers represent the aggregated scores across all fine-tuned tasks.
For aspect ratio~$> 1.0$, using Pix2Struct patching significantly outperforms a fixed grid patching, whereas for aspect ratio~$< 1.0$, a fixed grid patching outperforms Pix2Struct by a smaller margin.}
\label{fig:pix2struct-ablation}
\end{figure*}

\subsection{Ablation Studies}
In this section, we perform ablation studies evaluating (1) the impact of pix2struct patching and (2) using LLM generated data for pre-training. All ablation studies are performed on the 670M parameter variant.

\paragraph
{Impact of Pix2struct Patching.}
For this study, we compare a $670 \text{M}$ model using pix2struct patching with another using fixed-grid patching.
After pre-training, both models are fine-tuned on all tasks in Table~\ref{tab:finetuning_mixture}. We split each dataset into subsets based on the image aspect ratio and compute the respective metric on these subsets.
To compare fixed-grid patching to a variable pix2struct patching, we compute an \emph{aggregate score}, by first dividing the score of each task subset using fixed-grid patching by the score of the model using pix2struct on the entire task, and finally compute the geometric mean across all tasks.
Figure~\ref{fig:pix2struct-ablation} shows that for images with aspect ratio~$> 1.0$ (landscape mode images), the pix2struct patching strategy is significantly better than the fixed grid patching. For portrait mode images, the trend is reversed, but fixed grid patching is only marginally better.
Given that we want the ScreenAI model to be used across images of different aspect ratios, we choose to use pix2struct patching.

\paragraph{Impact of LLM Generated Data.}
For this experiment, we compare a $670 \text{M}$ ScreenAI model pre-trained using all the datasets mentioned in Section~\ref{sec:pretraining-mixtures} against a model pre-trained on a mixture excluding any LLM generated pre-training data. After pre-training, both models are fine-tuned on all tasks mentioned in Table~\ref{tab:finetuning_mixture} and an aggregate score is computed. We observe that adding LLM generated data to the mixture improves the aggregate score by $4.6$ percentage points.

\section{Conclusions}
In this work, we introduce the ScreenAI model along with a new unified schema for representing complex data and visual information, compatible with infographics, document images, and various UIs.
This unified representation enables the design of a mixture of self-supervised learning tasks, leveraging data from all these domains.
We show that training on this mixture results in a positive transfer to screen-related tasks as well as infographics and document-related tasks.
We also illustrate the impact of data generation using LLMs and justify our model design choices with ablation studies.
We apply these techniques to train a model that performs competitively and achieves SoTA on a number of public benchmarks. 
While our model is best-in-class, we note that, on some tasks, further research is needed to bridge the gap with models like GPT-4 and Gemini, which are orders of magnitude larger.
To encourage further research, we release a dataset with this unified representation, as well as two other datasets to enable more comprehensive benchmarking of models on screen-related tasks.

\section*{Acknowledgements}

We would like to thank team alumni Yo Hsiao and Zixian Ma for their contribution to the project, Fangyu Liu, Xi Chen, Efi Kokiopoulou, Jesse Berent, Gabriel Barcik, Lukas Zilka, Oriana Riva, Gang Li, Yang Li, Radu Soricut and Tania Bedrax-Weiss for their insightful feedbacks and fruitfull discussions, Rahul Aralikatte, Hao Cheng and Daniel Kim for their whole-hearted and tireless support in data preparation, and Jay Yagnik, Blaise Aguera y Arcas, Ewa Dominowska, David Petrou, and Matt Sharifi for their vision and support in leadership.

\section*{Contribution Statement}

\paragraph{First Authors with Equal Contributions:} Gilles Baechler, Srinivas Sunkara, Maria Wang, Jindong Chen.
\paragraph{Project Leads:} Jindong Chen, Abhanshu Sharma

\small
\bibliographystyle{named}
\bibliography{bibliography}

\appendix
\section*{Appendix}
\section{Definitions of Metrics}
\label{app:metrics}
We describe below the two categories of metrics that we use in our fine-tuning benchmarks.

\paragraph{Metrics for object detection tasks.}
For tasks involving the predictions of bounding boxes (UI elements), we use the standard object detection approach, which consists of first matching the predicted bounding boxes with the ground truth, and then computing various metrics from these matches. We set the Intersection over Union (IoU) threshold to $0.1$, and we perform the matching per class, not globally.
The metrics used in this paper are:
\begin{enumerate}
    \item \textbf{F1@IoU=0.1} - F1 score (harmonic mean of the precision and recall) at IoU threshold $0.1$.
    \item \textbf{Acc@IoU=0.1} - Top-1 accuracy at IoU threshold $0.1$.
\end{enumerate}

\paragraph{Metrics for benchmarks where output is plain text.}
For all other tasks, we use the following metrics:
\begin{enumerate}
    \item \textbf{CIDEr} - Consensus-based Image Description Evaluation~\protect\cite{vedantam2015cider}.
    \item \textbf{SQuAD F1} - F1 score (harmonic mean of the precision and recall) after applying SQuAD (Stanford Question Answering Dataset)~\protect\cite{rajpurkar2016squad} text pre-processing.
    \item \textbf{Relaxed accuracy}~\protect\cite{methani2020plotqa},
    \item \textbf{ANLS} - Average Normalized Levenshtein Similarity~\protect\cite{mathew2021docvqa}.
    \item \textbf{Exact Match(EM)} - See \url{https://github.com/huggingface/datasets/tree/main/metrics/exact_match#readme} for definition of Exact Match.
\end{enumerate}

\section{Screen Schema Examples}
\label{app:screen_schema}

\begin{figure}
\includegraphics[width=0.5\textwidth]{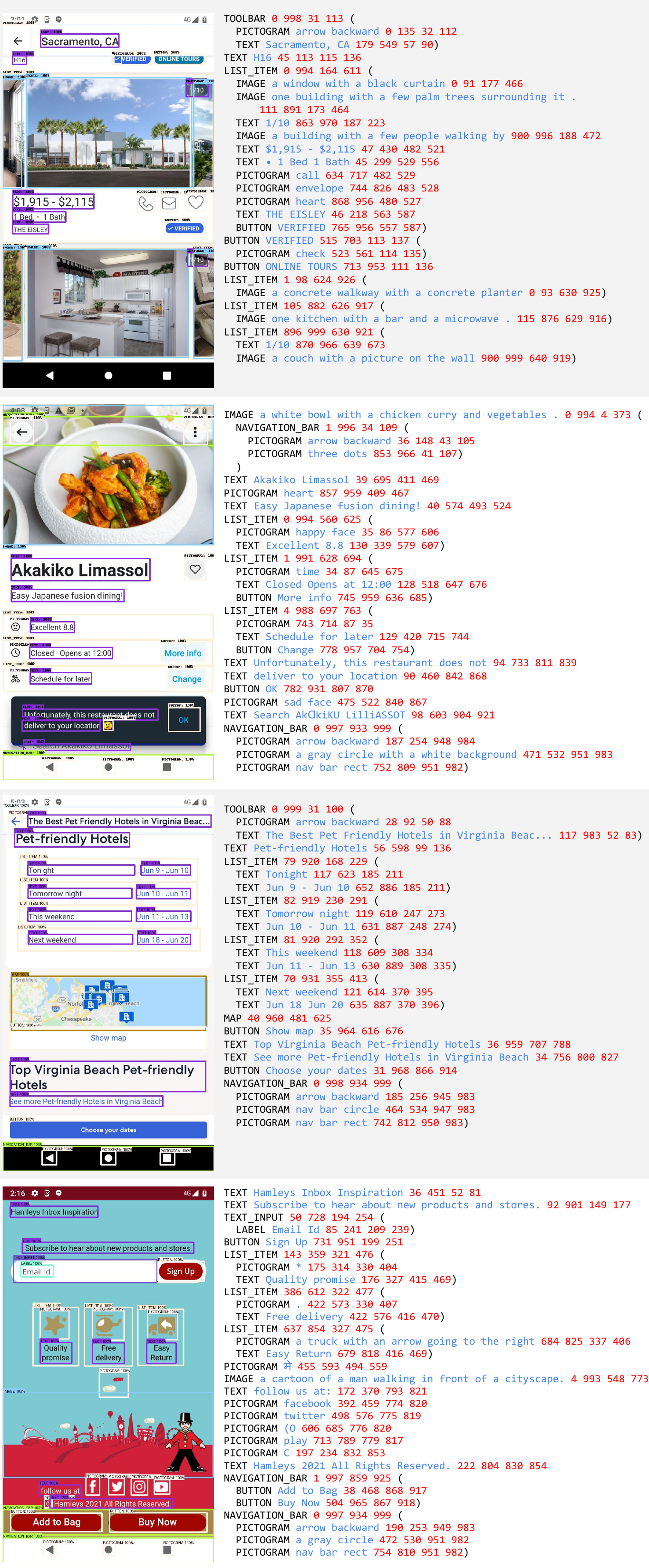}
\caption{Examples of our screen schema.}
\label{fig:screen_schema}
\end{figure}

Figure~\ref{fig:screen_schema} shows examples of the screen schema used in most of our pretraining tasks. Each schema contains:
\begin{enumerate}
    \item The UI element names.
    \item The OCR text (when applicable).
    \item The element descriptions (e.g. captioning, or the icon name).
    \item The bounding box coordinates, quantized and normalized between $0$ and $999$.
\end{enumerate}
Parentheses are used to create a basic hierarchical structure between the elements, i.e. the children of a parent element are all put inside a parenthesis block. For ease of visualization, the bounding boxes from the screen schema have been overlaid on the original screenshot.

\section{Prompts For LLM Generated Content}
\label{app:qa_example_prompt}

In this section, we present some of the prompts used as input to LLMs like PaLM 2-S~\cite{anil2023palm} to generate data for screen question answering, screen navigation and screen summarization tasks.
In addition to the prompt, we also pass as input to the LLM the screen annotation schema described in Appendix~\ref{app:screen_schema}.

\subsection{Screen Question Answering}
\begin{itemize}
\item []
\begin{lstlisting}
You only speak JSON. Do not write text that isn't JSON. 
You are given the following mobile screenshot, described in words. Can you generate 5 questions regarding the content of the screenshot as well as the corresponding short answers to them? The answer should be as short as possible, containing only the necessary information. Your answer should be structured as follows:
questions: [
{{question: the question,
 answer: the answer
}}, ...]
{THE SCREEN SCHEMA}
\end{lstlisting}
\end{itemize}

\subsection{Screen Navigation}
\begin{itemize}
\item []
\begin{lstlisting}
You only speak JSON. Do not write text that isn't JSON. You are given a mobile screenshot, described in words. Each UI element has a class, which is expressed in capital letter. The class is sometimes followed by a description, and then 4 numbers between 0 and 999 represent the quantized coordinates of each element.
Generate {num_samples} single-step navigation instructions and their corresponding answers based on the screenshot. Each answer should always start with `click`, followed by the coordinates of the element to click on, e.g. `click 0 137 31 113`.
Be creative with the questions, do not always use the same wording, refer to the UI elements only indirectly, and use imperative tense. Your answer should be structured as in the example below:

"questions": [
{{"question": "the question",
  "answer": "click 0 137 31 113"
}},
...
]
{THE SCREEN SCHEMA}
\end{lstlisting}
\end{itemize}

\subsection{Screen Summarization}
\begin{itemize}
\item []
\begin{lstlisting}
You only speak JSON. Do not write text that isn't JSON.
You are given the following mobile screenshot, described in words.
Generate a summary of the screenshot in 2-3 sentences. Do not focus on specifically naming the various UI elements, but instead, focus on the content. Your answer should be structured as follows:
"summary": the screen summary
{THE SCREEN SCHEMA}
\end{lstlisting}
\end{itemize}

\section{Screen Navigation Generated Examples}
\label{appendix:screen_navigation_generated_examples}
\begin{figure}
\includegraphics[width=0.5\textwidth]{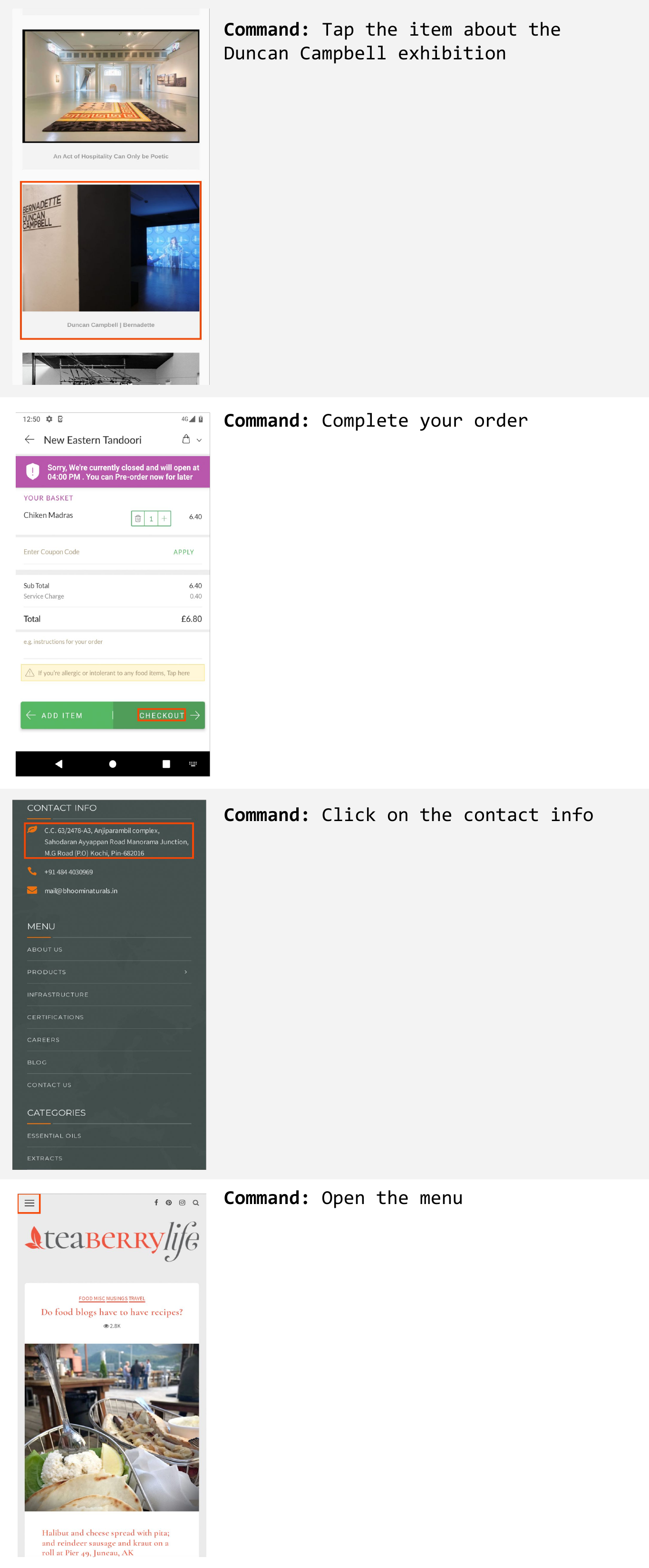}
\caption{Examples of Screen Navigation data generated using an LLM. The target bounding box is highlighted in red.}
\label{fig:screen_nav_generated}
\end{figure}

We present a few examples for the Screen Navigation task generated using LLMs in Figure~\ref{fig:screen_nav_generated}. More details about the data generation process can be found in Section~\ref{sec:automatic-data-gen}.

\section{MoTIF Evaluation Results}
\label{appendix:motif_eval_results}

\begin{table}[t]
\centering
\begin{tabular}{ccc}
\toprule
\textbf{Model} & \textbf{App Seen}  & \textbf{App Unseen} \\
\midrule
Baseline & 66.3 & 67.6 \\
ScreenAI & \textbf{87.7} & \textbf{87.8} \\
\bottomrule
\end{tabular}
\caption{Metrics on different splits of 
MoTIF~\protect\cite{burns2022motif}
Task Automation.}
\label{tab:motif_results_deails}
\end{table}

In this section, we present the ScreenAI model metrics on the different splits of the MoTIF~\cite{burns2022motif} task automation dataset. The metrics breakdown can be seen in Table~\ref{tab:motif_results_deails}.

\section{ScreenQA Short Answers Generation}
\label{appendix:screenqa_short}

\begin{figure}
\includegraphics[width=0.5\textwidth]{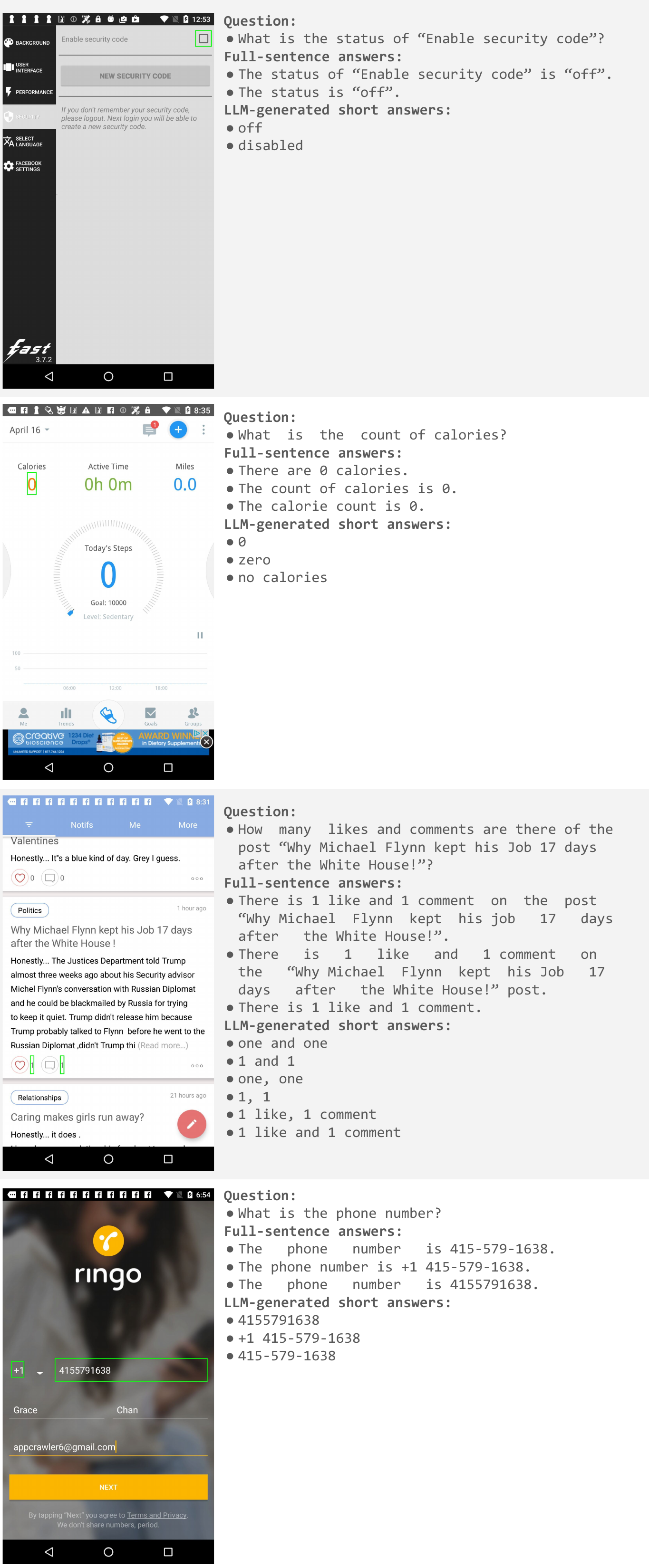}
\caption{Examples of questions and answers from the ScreenQA dataset, together with their LLM-generated short answers.}
\label{fig:screenqa_short_examples}
\end{figure}

We describe below the motivation behind producing a list instead of a single short answer as a new ground truth for the ScreenQA~\cite{hsiao2022screenqa} dataset, as well as the generation details.

There are many ways to represent the same information.
For example, "25.01.2023", "25th of January 2023" and "January 25, 2023" are representing the same date, and the model should not be penalized for choosing one representation over the others.
A list of various representations of the same factual answer allows this.

A variant of the PaLM 2-S~\cite{anil2023palm} was used to generate this list of short answers in a few-shot setting.
We give as input to the LLM text information from the ScreenQA dataset (question, list of UI elements descriptions and full-sentence answer) in addition to the prompts described in Appedix~\ref{appendix:screenqa_short_prompt_single} and \ref{appendix:screenqa_short_prompt_multiple}.
The generated lists were then verified by simple heuristics and eyeballing of random samples.
See examples of questions and answers from the ScreenQA task, together with their LLM-generated short answers, in Figure~\ref{fig:screenqa_short_examples}.

\subsection{For answers contained in a single UI element}
\label{appendix:screenqa_short_prompt_single}

For each entry in the ScreenQA dataset where there is \textbf{only one} UI element in the ground truth, we use the following prompt with the PaLM 2-S model~\cite{anil2023palm} to generate a list of short answers from the question, list of elements, and the full-sentence answer:

\lstset{literate=
    {°}{{$^\circ$}}1
    {®}{{\textsuperscript{\textregistered}}}1
    {™}{{\texttrademark}}1
    {⇆}{{$\rightleftarrows$ }}2
}
\begin{lstlisting}
List various ways to rephrase the answer. The answer should be as short as possible, without extra words from the question. Use all provided elements in each answer. Provide the output in square brackets.

Here is an example:
Question: 'What's the percentage of humidity?'
Answer elements: ['65%
Full answer: 'The humidity is 65%
Rephrases: ['65%

Here is another example:
Question: 'What is the gender?'
Answer elements: ['Male']
Full answer: 'The gender is male.'
Rephrases: ['male']

Here is another example:
Question: 'What is the status of "24 hr clock"?'
Answer elements: ['on']
Full answer: 'The status is "on".'
Rephrases: ['on', 'enabled']

[...]

Now is your turn.
Question: {THE QUESTION}
Answer elements: {THE UI ELEMENT DESCRIPTION}
Full answer: {THE FULL-SENTENCE ANSWER}
Rephrases:
\end{lstlisting}

\subsection{For answers contained in multiple UI elements}
\label{appendix:screenqa_short_prompt_multiple}

For each entry in the ScreenQA dataset where there are \textbf{more than one} UI elements in the ground truth, we use the following prompt with the PaLM 2-S model~\cite{anil2023palm} to generate a list of short answers from the question, list of UI elements and full-sentence answer:

\begin{lstlisting}
List various ways to rephrase the answer. The answer should be as short as possible, without extra words from the question. Use all provided elements in each answer. Provide the output in square brackets.

Here is an example:
Question: 'What's the temperature?'
Answer elements: ['59', '°F']
Full answer: 'The temperature is 59 degrees Fahrenheit.'
Rephrases: ['59°F', '59 Fahrenheits', '59 degrees Fahrenheit']

Here is another example:
Question: 'What is the name?'
Answer elements: ['Jon', 'Brown']
Full answer: 'The name is Jon Brown.'
Rephrases: ['Jon Brown']

Here is another example:
Question: 'What is the rest interval duration?'
Answer elements: ['00', ':', '34']
Full answer: 'The rest interval lasts 00:34.'
Rephrases: ['00:34', '34 seconds', '0 minutes and 34 seconds', '34 minutes', '0 hours and 34 minutes']

[...]

Now is your turn.
Question: {THE QUESTION}
Answer elements: {THE FIRST UI ELEMENT DESCRIPTION, ...}
Full answer: {THE FULL-SENTENCE ANSWER}
Rephrases:
\end{lstlisting}

\section{Complex Question Answering Datasets}
\label{appendix:complex_qa_datasets}

\begin{figure}
\includegraphics[width=0.5\textwidth]{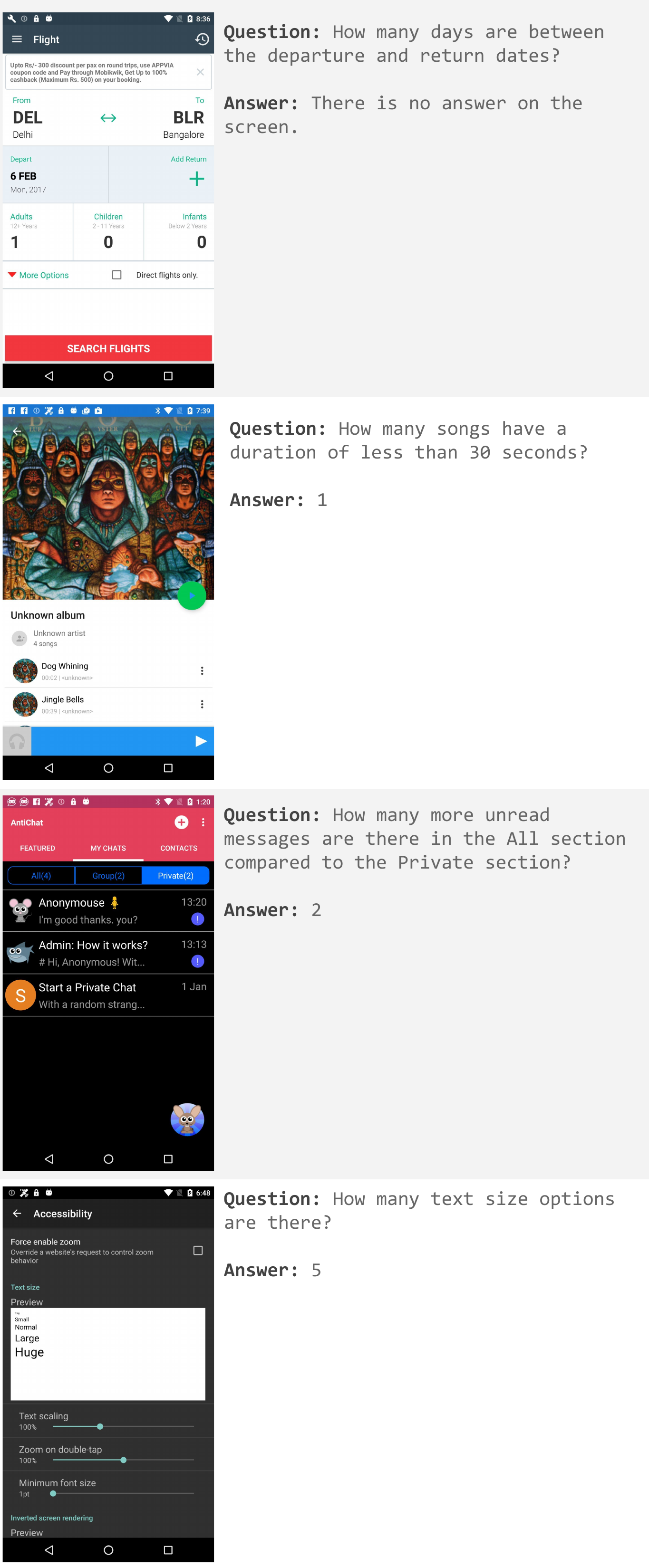}
\caption{Examples of mobile Complex QA evaluation examples.}
\label{fig:complex_mobile}
\end{figure}

\begin{figure}
\includegraphics[width=0.5\textwidth]{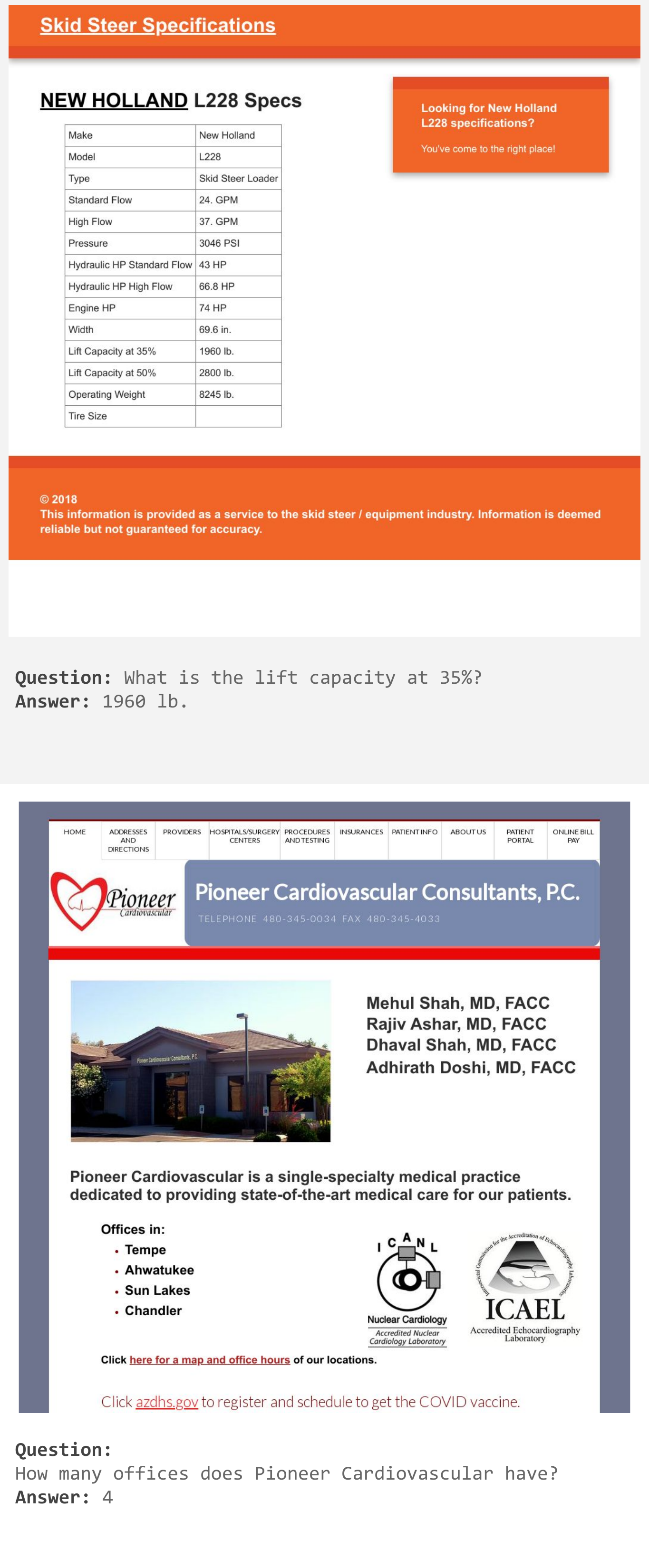}
\caption{Examples of desktop Complex QA evaluation examples.}
\label{fig:complex_desktop}
\end{figure}

The Complex QA datasets contain machine-generated questions using LLMs like PaLM 2-S~\cite{anil2023palm} based on the Screen Annotation output from the best ScreenAI VLM.
For each dataset, the prompts are chosen to target certain types of questions.
With this approach, we generate large scale datasets for desktop, mobile, mobile with different aspect ratios, and infographics screens.
These datasets are used both for pre-training and evaluation.
We add an additional step of human raters verification for the evaluation data.
Figure~\ref{fig:complex_mobile} and Figure~\ref{fig:complex_desktop} show a few examples of LLM-generated QA data that was verified by humans.

We distinguish three different subsets, each focusing on solving the various challenges we identified with this task:

\begin{itemize}
    \item \textbf{Desktop QA and Long Webpage QA}: Datasets on desktop screens and long (viewport height) webpages, respectively. The aspect ratio and size of the input images is very different compared to other QA datasets.

    \item \textbf{Complex QA datasets}: Datasets mainly focused on counting, arithmetic, and comparison operations requiring information from more than one part of the screen. 
    \begin{itemize}
        \item Complex QA: Mobile app screens
        \item Desktop Complex QA: Desktop screens.
        \item Long Webpage Complex QA: Long webpages.
    \end{itemize}

    \item \textbf{Non Answerable QA}: Dataset focused on measuring the ability of the model to know when a question cannot be answered from the given screen.
\end{itemize}

\section{New Benchmarks Repositories}

We release three evaluation datasets for tasks described in Section~\ref{sec:fine-tuning-mixtures}:
\begin{itemize}
    \item
    \textbf{Screen Annotation (SA):}
    \url{https://github.com/google-research-datasets/screen_annotation}
    \item
    \textbf{ScreenQA Short (SQA Short):} 
    \url{https://github.com/google-research-datasets/screen_qa?tab=readme-ov-file#screenqa-short}
    \item
    \textbf{Complex ScreenQA (Cplx SQA): }
    \url{https://github.com/google-research-datasets/screen_qa?tab=readme-ov-file#complexqa}
\end{itemize}

\end{document}